\documentclass[a4paper]{cas-sc}

\usepackage[authoryear]{natbib}
\usepackage{amsmath,amssymb,amsfonts}
\usepackage{algorithmic}
\usepackage{algorithm}
\usepackage{array}
\usepackage[caption=false,font=normalsize,labelfont=sf,textfont=sf]{subfig}
\usepackage{textcomp}
\usepackage{stfloats}
\usepackage{url}
\usepackage{verbatim}
\usepackage{graphicx}
\usepackage{xcolor}
\usepackage{booktabs}
\usepackage{tabularray}
\usepackage{color}
\definecolor{Mercury}{rgb}{0.901,0.901,0.901}

\def\tsc#1{\csdef{#1}{\textsc{\lowercase{#1}}\xspace}}
\tsc{WGM}
\tsc{QE}
\tsc{EP}
\tsc{PMS}
\tsc{BEC}
\tsc{DE}

\begin{document}
\let\WriteBookmarks\relax
\def\floatpagepagefraction{1}
\def\textpagefraction{.001}
\shorttitle{Mixture-of-Modality-Experts with Holistic Token Learning for Fine-Grained Multimodal Visual Analytics}
\shortauthors{Tianyi Liu et~al.}

\title [mode = title]{Mixture-of-Modality-Experts with Holistic Token Learning for Fine-Grained Multimodal Visual Analytics in Driver Action Recognition}

\author[1]{Tianyi Liu}[orcid=0000-0002-6705-7808]
\ead{liut0038@e.ntu.edu.sg}
\credit{Conceptualization, Methodology, Software, Data curation, Formal analysis, Validation, Visualization, Investigation, Writing - original draft, Project administration}
\affiliation[1]{organization={Nanyang Technological University},
                addressline={50 Nanyang Avenue}, 
                postcode={639798}, 
                country={Singapore}}

\author[1]{Yiming Li}[orcid=0000-0001-5871-2316]
\ead{yiming.li@ntu.edu.sg}
\credit{Conceptualization, Methodology, Data curation}

\author[1,2]{Wenqian Wang}
\fnmark[1]
\ead{wenqian_wang@sutd.edu.sg}
\credit{Conceptualization, Methodology, Software}
\affiliation[2]{organization={Singapore University of Technology and Design},
                addressline={8 Somapah Rd}, 
                postcode={487372}, 
                country={Singapore}}

\author[1]{Jiaojiao Wang}[orcid=0000-0002-2399-5296]
\ead{jiaojiao001@e.ntu.edu.sg}
\credit{Formal analysis, Visualization, Writing - review \& editing}

\author[1]{Chen Cai}[orcid=0009-0002-7793-5261]
\ead{e190210@e.ntu.edu.sg}
\credit{Writing - review \& editing, Investigation}

\author[3]{Yi Wang}[orcid=0000-0001-8659-4724]
\ead{yi-eie.wang@polyu.edu.hk}
\credit{Writing - review \& editing, Investigation}
\affiliation[3]{organization={The Hong Kong Polytechnic University},
                addressline={6/F, Core D, Lui Che Woo Building, Hung Hom},
                city={Hong Kong SAR},
                country={China}}

\author[1]{Kim-Hui Yap}[orcid=0000-0003-1933-4986]
\cormark[1]
\ead{ekhyap@ntu.edu.sg}
\credit{Supervision, Resources, Funding acquisition, Project Administration}

\cortext[cor1]{Corresponding author}
\fntext[fn1]{Wenqian Wang was with NTU when this research was conducted and is currently with SUTD.}

\begin{abstract}
Robust multimodal visual analytics remains challenging when heterogeneous modalities provide complementary but input-dependent evidence for decision-making.
Existing multimodal learning methods mainly rely on fixed fusion modules or predefined cross-modal interactions, which are often insufficient to adapt to changing modality reliability and to capture fine-grained action cues. 
To address this issue, we propose a Mixture-of-Modality-Experts (MoME) framework with a Holistic Token Learning (HTL) strategy. 
MoME enables adaptive collaboration among modality-specific experts, while HTL improves both intra-expert refinement and inter-expert knowledge transfer through class tokens and spatio-temporal tokens. 
In this way, our method forms a knowledge-centric multimodal learning framework that improves expert specialization while reducing ambiguity in multimodal fusion.
We validate the proposed framework on driver action recognition as a representative multimodal understanding task
The experimental results on the public benchmark show that the proposed MoME framework and the HTL strategy jointly outperform representative single-modal and multimodal baselines. 
Additional ablation, validation, and visualization results further verify that the proposed HTL strategy improves subtle multimodal understanding and offers better interpretability.
\end{abstract}




\maketitle

\section{Introduction}
Multimodal visual analytics systems are increasingly required to make reliable decisions from heterogeneous evidence whose informativeness may vary across inputs and environments. 
In such settings, robust multimodal understanding depends not only on combining complementary modalities but also on adaptively organizing their contributions according to the current observation.
This challenge is particularly evident in driver action recognition (DAR). 
In smart car cabins, DAR is a representative multimodal visual analytics problem, where unpredictable variations in illumination and sensor modalities pose unique challenges to system robustness~\citep{kuang2023mifi}. 
Unlike general action recognition, in-cabin actions usually involve limited body movement, subtle local motion, and a largely fixed background.
Meanwhile, illumination variation, occlusion, and sensing heterogeneity further reduce the reliability of any single modality. 
As a result, robust DAR requires not only visual perception, but also effective integration of complementary knowledge from multiple modalities.
However, multimodal integration in DAR is inherently challenging because the usefulness of modality-specific knowledge is input-dependent.

Existing multimodal DAR methods, and many multimodal recognition models in general, mainly improve performance through feature fusion~\citep{lin2024dfs, wang2024cm2} or predefined cross-modal interaction~\citep{wang2024multifuser}. 
However, such strategies are typically controlled by fixed learned parameters and may become inadequate when the reliability of the modality varies between samples and types of action. 
This limitation suggests that multimodal DAR should be formulated not merely as feature aggregation, but as adaptive collaboration among specialized modality experts that encode different knowledge sources, so that heterogeneous evidence can be dynamically organized for more robust decision-making.

However, adaptive expert collaboration alone is still insufficient for fine-grained multimodal understanding. 
In in-cabin environments, a central challenge is that large-scale pre-trained transformer backbones may rely on coarse contextual patterns when the scene is visually constrained and largely static, thereby overlooking subtle action-relevant motion cues~\citep{chung2022backgroundbias}. 
This issue is particularly critical for DAR, where many action categories are distinguished by fine-grained spatio-temporal details. 
Therefore, beyond dynamic inter-expert collaboration, each expert should also refine how action-relevant knowledge is represented and propagated internally. 
This motivates a holistic token learning strategy that treats class tokens and spatio-temporal tokens as complementary carriers of multimodal action knowledge. 
In such strategy, inter-expert knowledge transfer supports high-level coordination, while intra-expert self-guidance further refines both class-token semantics and spatio-temporal token representations for finer-grained action perception.

To address the mentioned problems, we propose a novel DAR framework named \textbf{Mixture-of-Modality-Experts (MoME)}, which is designed to comprehensively model driver actions across different visual modalities through specialized modality-specific experts.
Compared with previous methods that rely on fixed-weight coordination among experts, MoME adopt a dynamic gating mechanism, enabling flexible collaboration among modality experts and establishing a Mixture-of-Experts (MoE)-enhanced DAR system.
To better achieve fine-grained analysis of subtle motion patterns and mitigate background bias, we further propose a \textbf{Holistic Token Learning (HTL)} strategy.
HTL is specifically designed to address several limitations in existing research, thereby improving how MoE-based multimodal models coordinate, refine, and transfer fine-grained action knowledge in constrained environments:
(i) While patch-level supervision has been explored for vision transformers~\citep{jiang2021tokenlabelling}, the role of spatio-temporal tokens in video transformer architectures remains largely unexplored;
(ii) Existing distillation methods for vision transformers are limited in scope: self-distillation typically focuses only on class tokens~\citep{sultana2022self}, while inter-expert distillation, as in the Mixture-of-Distilled-Experts (MoDE) framework~\citep{xie2024mode}, is also restricted to class token interactions without considering intra-expert knowledge transfer or spatio-temporal token modeling.
HTL fills the gap mentioned above, it performs comprehensive token-level distillation by simultaneously refining both class tokens and spatio-temporal tokens within and among each expert.
This design enables both inter- and intra-expert knowledge transfer, promoting fine-grained feature refinement and improving the robustness of MoE-enhanced DAR systems in constrained in-cabin environments.

In summary, our contribution can be summarized as follows: 
\begin{itemize}
    \item We propose Mixture-of-Modality-Experts (MoME), a knowledge-centric multimodal framework, which first investigates the effectiveness of MoE architecture in enabling flexible collaboration across different sensing modalities. 
    
    \item We introduce Holistic Token Learning (HTL) mechanism to facilitate effective inter-modality knowledge transfer and intra-modality self-refinement. Unlike traditional expert collaboration schemes, HTL enables each modality expert to refine their action recognition capabilities by distilling knowledge from both peer experts and itself, thereby achieving more robust and finer-grained multimodal visual analytics.
    
    \item Extensive experiments show that the proposed MoME framework and HTL mechanism jointly contribute to the state-of-the-art multimodal visual analytics performance in DAR task. Additionally, improved interpretation can also be obtained.

\end{itemize}

\begin{figure}
    \centering
    \includegraphics[width=0.95\linewidth]{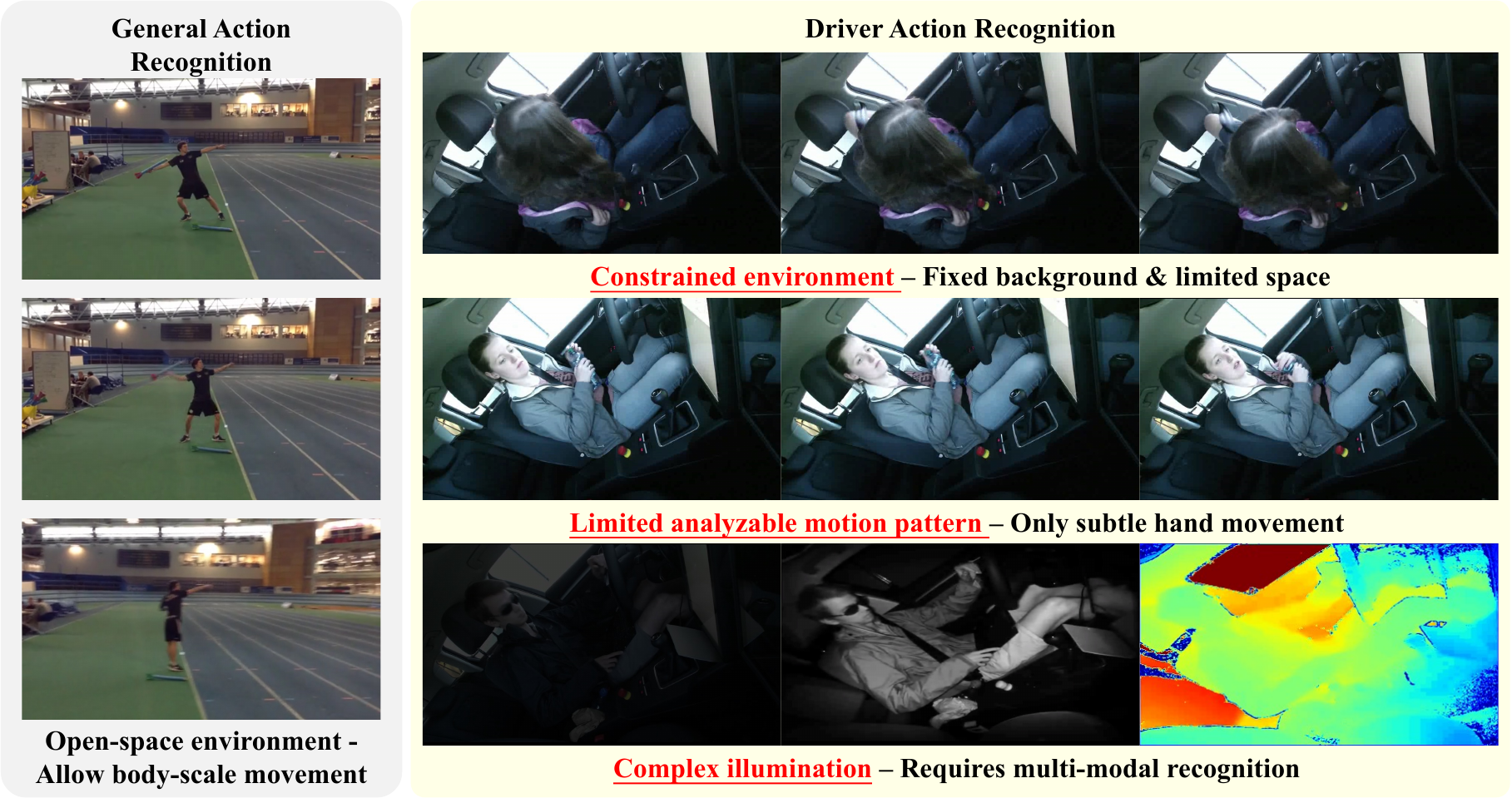}
    \caption{In-cabin action recognition presents subtle motion pattern and complicated illumination variation, making fine-grained multimodal evidence sparse and unevenly reliable. These challenges therefore requires robust multimodal visual analytics to adaptively integrate knowledge from different sources.}
    \label{fig:motivation}
\end{figure}

\section{Related Works}
In this section, we will briefly review related research on driver activity monitoring, including traditional driver fatigue and distraction detection, as well as comprehensive action recognition. 

\subsection{Traditional In-Cabin Monitoring} 
Unlike traditional recognition tasks~\cite{li2023uniformer, li2023uniformerv2, liu2022interactive}, traditional driver activity monitoring focuses mainly on driving safety issues such as driver fatigue detection and distraction recognition based on physiological signal processing such as eye movement, head position, and EEG signal~\citep{lv2022compact, gong2024tfac}, as fatigue and distraction are a key factor in traffic accidents.
In recent years, vision-based driver fatigue and distraction detection shows more guarantee performance and is easy to deploy without invasion.
Ma et al. proposed to leverage depth~\citep{ma2017depth} and infrared~\citep{ma2019conv} video to identify driver fatigue using multi-stream fusion approaches.
Yang et al. proposed a 3D deep learning network with a low time sampling rate that uses subtle facial action recognition information to guide the detection of driver yawness~\citep{yang2020yawn}.
Kuang et al. proposed the MultI-Camera Feature Integration (MIFI) approach for 3D distracted driver action recognition which utilize various multi-camera feature fusion strategies and a special training objective that reweighs easy and hard samples~\citep{kuang2023mifi}.

\subsection{Driver Action Recognition} 
However, these traditional driver monitoring researches have limited scope of safety management, in addition, current smart cabin designs are devoted to achieve comprehensive analytics of the in-cabin activities. 
Smart cabins must improve the interaction between humans and vehicles and enable intelligent action interventions, and recent developments have expanded these approaches. 
For example, after Martin et al. introduced the large-scale multimodal dataset Drive\&Act~\citep{martin2019daa}, Peng et al. integrated Swin-Transformer into driver action recognition, incorporating a feature calibration to improve model generalizability~\citep{peng2022transdarc}. 
In addition to Drive\&Act, Tan et al. proposed another dataset in the bus environment, using driver posture as complementary information; it achieves effective recognition~\citep{tan2021bidirectional}.
However, these approaches often continue to rely on single-modality inputs, neglecting the advantages offered by multimodal data.
Based on the idea of inter-modality interaction, Lin et al. proposed the dual feature shift approach with neighbor feature propagation for efficient multimodal driver action recognition~\citep{lin2024dfs} and Wang et al. proposed Multifuser which performs multimodal feature integration~\citep{wang2024multifuser}.
Furthermore, Wang et al. suggested the use of continuous learning to link instructive prompts with new modalities to improve training guidance~\citep{wang2024cm2}.

\subsection{Model Enhancement for Multimodal Collaboration}
As discussed above, existing DAR studies mainly rely on early-fusion, late-fusion, or dedicated cross-modal fusion modules to combine heterogeneous visual inputs. 
Although these strategies have demonstrated the benefit of multimodal learning, the fusion process is usually governed by fixed parameters learned during training. 
Such static fusion mechanisms are often insufficient for complex in-cabin scenarios, where the reliability and informativeness of different modalities may vary across samples, illumination conditions, and action categories. 
This limitation motivates the introduction of Mixture-of-Experts (MoE), which provides a more adaptive formulation for multimodal understanding by allowing different experts to contribute dynamically according to the input.

Beyond adaptive expert collaboration, improving knowledge transfer within the model has also become increasingly important for handling challenging recognition scenarios. 
For transformer backbones, Sultana et al.~\citep{sultana2022self} proposed self-distillation for vision transformers to improve domain generalization by transferring knowledge between different layers. 
However, their design mainly focuses on class-token-level supervision, which is more suitable for global semantic alignment than for capturing the subtle motion patterns required in DAR. 
For MoE structures, Xie et al.~\citep{xie2024mode} proposed MoDE, which enhances expert collaboration through mutual distillation among experts. 
This study suggests that expert interaction can improve the effectiveness of MoE-based models. 
Nevertheless, the interaction in MoDE is still mainly performed at the class-token level, and its validation is mainly conducted on general recognition benchmarks.
In addition, inspired by previous research in the video field~\cite{sultana2022self, liu2024bitstream, liu2025towards, liu2026accelerating}, fine-grained interactions at the spatio-temporal token level are worth further investigation for enhancing DAR models.
Therefore, it remains unclear whether these designs are beneficial for fine-grained multimodal understanding in practical DAR scenarios. 
These observations indicate that robust DAR requires not only adaptive modality-expert collaboration but also finer-grained token-level knowledge transfer beyond class-token-level supervision.

\section{Methodology}
In this section, we introduce our methodology in detail.
Specifically, we first introduce the proposed Mixture-of-Mondaility-Experts (MoME) framework and describe how we train MoME with the holistic token learning (HTL) strategy. 

\subsection{Problem Statement}~\label{problem}
The multimodal driver action recognition task aims to recognize driver behaviors using inputs from multiple sensor modalities equipped in the car cabin, such as RGB, infrared, depth data, etc.~\citep{wang2024multifuser}. 
Each modality provides complementary information that, when jointly learned, is expected to lead to a more robust and accurate action recognition model. 
Specifically, we consider a scenario with \textit{N} sychronized modalities. 
Each sample can be represented as \( \mathbf{x}_i = \{(\mathbf{x}^1_i, \mathbf{x}^2_i, ..., \mathbf{x}^N_i), \mathbf{y}_i\} \), where \( \mathbf{x}^n_i \in \mathbb{R}^{C_n \times T \times H \times W} \) denotes the input frame sequence from the $n$-th visual modalities. 
Specifically, here \( C_n \) represents the number of channels specific to the modality $n$, while \( T \), \( H \), and \( W \) correspond to the temporal length, height, and width of the input frame sequence, respectively. 
\( \mathbf{y}_i \) is the corresponding label for sample $\mathbf{x}_i$ in this multi-class recognition problem. 
The key challenge is not only to combine heterogeneous modality features but also to determine how much each modality should contribute under different observations. 
Because modality-specific signals can vary across appearance, noise sensing, and environmental conditions, their reliability is inherently input-dependent. 
This property makes static multimodal fusion insufficient and motivates an adaptive expert-based formulation for robust multimodal representation learning.

\begin{figure*}
    \centering
    \includegraphics[width=1\textwidth]{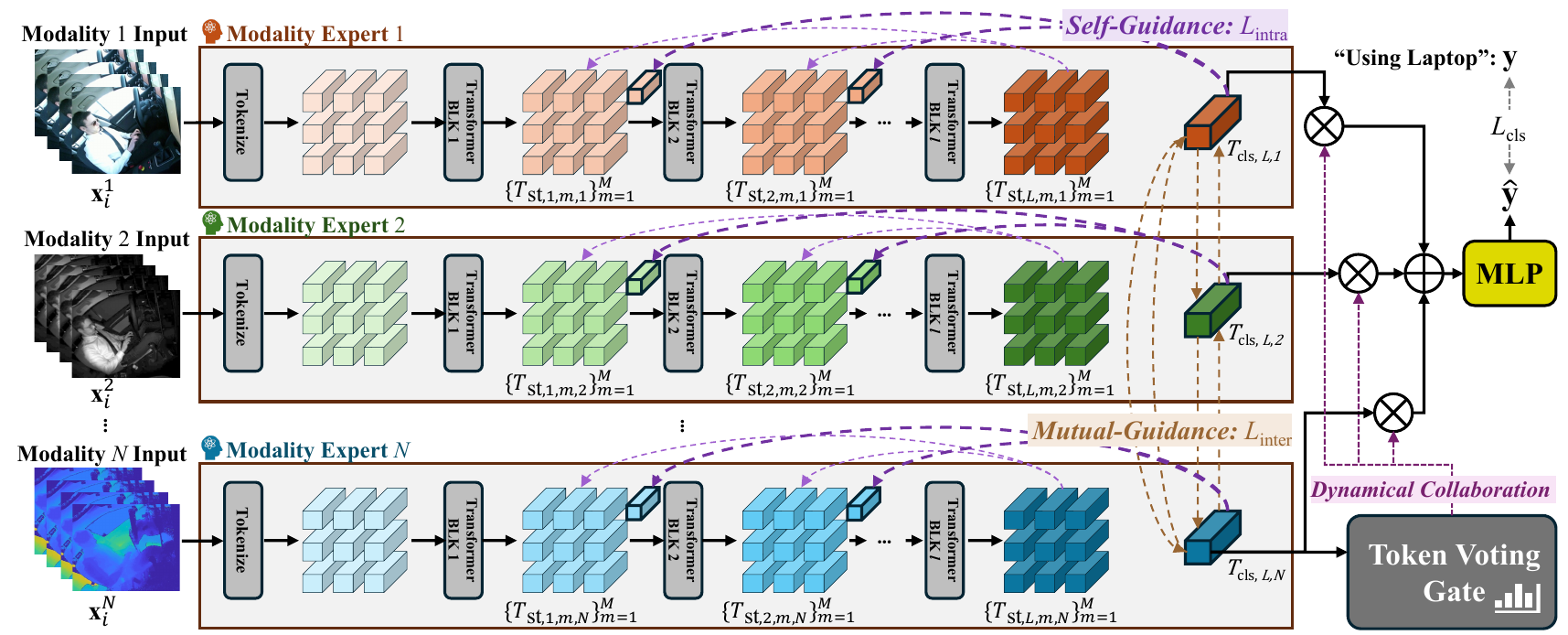}
    \caption{Overview of our proposed Mixture-of-Modality-Experts (MoME) framework enhanced by the Holistic Token Learning (HTL) strategy.}
    \label{fig:moe_overview}
\end{figure*}

\subsection{Mixture-of-Modality-Experts (MoME)}~\label{tl}
Unlike previous research that simply relies on fusion blocks with fixed fusion weights learned during training, we firstly introduce the idea of mixture-of-experts (MoE) to DAR task and build our Mixture-of-Modality-Experts (MoME) framework, which is able to dynamically generates input-dependent weights through a gating mechanism, enabling flexible and adaptive fusion.
The concept of MoE was first introduced by Jacobs et al.~\citep{jacobs1991moe} to combine various experts, each learned from distinct data, into a single robust model. 
In the MoE setup, a voting gate is used as the common gating mechanism that assigns weights to each expert based on the input, determining how much influence each expert has on the final prediction. 
Traditionally, the gating module receives the input sample $\mathbf{x}_i$ and generate a normalized weighting vector denoted by $\text{SoftMax}(g_n(\mathbf{x}_i))$, which corresponds to as set of experts output denoted by $\text{E}_n(\mathbf{x}^n_i)$.

In MoME, different from conventional MoE formulations that generate routing weights solely from input features, our gating module is conditioned on expert outputs, i.e., the class tokens $T_{\text{cls}, L, n} = \text{E}_n(\mathbf{x}^n_i)$ produced by modality experts. 
This output-aware design is motivated by the observation that decision-oriented representations often encode richer evidence about expert confidence and action-specific response patterns than raw inputs alone. 
As a result, the gate can make more goal-aligned coordination decisions by directly examining the predictive status of each expert before fusion.
The final latent embedding $\mathbf{z}$ is a weighted combination of the outputs of all experts:
\[
\mathbf{z} = \sum_{n=1}^{N}T_{\text{cls}, n} \cdot \left[ \text{Gate}([T_{\text{cls}, L, 1}; T_{\text{cls}, L, 2};\ldots ;T_{\text{cls}, L, N}]) \right]_n
\]
where the input of the gating module $\text{Gate}(\cdot)$ is the stacked output of the $N$ modality experts and $\text{SoftMax}(\cdot)$ is applied to normalize the weights to achieve reasonable enhancement and suppression among the experts. 
The voting gate thus serves as an adaptive coordination mechanism, distributing the decision-making process across different modalities rather than concentrating it on a single one.
The resulting embedding $\mathbf{z}$ is finally input to an MLP-based classifier $\mathcal{C}(\cdot)$ to predict the final output denoted as $\hat{\mathbf{y}} = \mathcal{C}(\mathbf{z})$.

\subsection{Holistic Token Learning}
While MoME improves multimodal coordination at the expert level, robust fine-grained recognition also requires each expert to refine how action-relevant knowledge is represented internally. 
To this end, we propose Holistic Token Learning (HTL), a unified token-level learning strategy that organizes knowledge transfer at two complementary levels: intra-expert refinement and inter-expert collaboration. 
Although prior studies on this topic remain limited, Mixture-of-Distilled-Experts (MoDE), a foundational work proposed in 2024 was designed to enhance expert interaction within MoE frameworks to balance individual contributions and improve the overall recognition performance.
However, MoDE primarily focuses on high-level class token information, overlooking the spatio-temporal token representations and their interrelations modeled by video transformers.
Moreover, MoDE has yet to demonstrate its effectiveness on downstream tasks.
Unlike prior methods that mainly focus on class-token supervision, HTL jointly models class tokens and spatio-temporal tokens, enabling finer-grained action knowledge to be refined and propagated throughout the multimodal expert system.

HTL fully leverages all output tokens of the video transformer, including both intermediate and output spatio-temporal tokens and the class tokens, thereby providing multi-level knowledge for action interpretation.
Specifically, as shown in Fig.~\ref{fig:moe_overview}, for intra-expert learning, HTL performs self-guidance in a deep-to-shallow manner. 
The deeper representations, which encode more mature semantic understanding, are used to regularize shallower spatio-temporal tokens. 
This process encourages earlier layers to better preserve subtle action-relevant motion cues, reduces over-reliance on coarse contextual patterns, and improves the internal consistency of expert representations.
For inter-expert learning, HTL introduces mutual guidance among modality experts. 
Since different experts observe the same action from heterogeneous sensing spaces, their representations contain complementary but not identical evidence. 
HTL therefore encourages semantic alignment at the expert-output level, allowing reliable decision knowledge to be transferred across modalities while preserving expert specialization. 
This design improves collective consistency without collapsing modality diversity.
Through this two-stage enhancement including self-guided low-level modeling and mutual high-level guidance, our MoME design benefits both individual expert robustness and stronger collective interaction, ultimately leading to improved overall performance.

In detail, the knowledge transfer is realized by setting auxiliary training objective. 
In general image recognition task, the training model can benefit from a large pre-trained model by referring to the patch tokens' denoted score map~\citep{jiang2021tokenlabelling}.
Here we extend this idea into a self-supervised spatio-temporal domain since DAR is a downstream video task lacks large pre-trained knowledge source.
To be specific, given a the input sample $\mathbf{x}_i$, denote the set of expert outputs of their last transformer blocks as $[T_{\text{cls}, 1}, ..., T_{\text{cls}, N}, T_{\text{st},1,1,1}, ..., T_{\text{st},L,M,N}]$, where $L$ denotes the total number of spatio-temporal tokens in each monolithic transformer blocks in each expert, $M$ is the number of transformer blocks in each expert, $N$, as mentioned above, the the number of modality experts.
$T_{\text{cls}, n}$ and $T_{\text{st},l,m,n}$ correspond to the class token and the spatio-temporal tokens, respectively. 
Mathematically, the basic classification loss for the sample $\mathbf{x}_i$ can be written as
$L_{\text{cls}} =H(\hat{\mathbf{y}}, \mathbf{y})$,
where $H(\cdot, \cdot)$ is the softmax cross-entropy loss and $\mathbf{y}$ is the ground-truth label.
In HTL, we use KL divergence, denoted as $D(\cdot || \cdot)$, for intra-expert self-guidance because tokens within the same expert share a consistent representational space, making distribution-level alignment appropriate,
\[
L_{\text{intra}} = \frac{\sum_{n=1}^{N} \sum_{l=1}^{L-1} \mathbf{w}_{l} (D(P_{L,n} || P_{l,n}) + D(Q_{L,n} || Q_{l,n}))}{N(L-1)} ,
\]
where $\mathbf{w}_{l}$ represents the learnable block-wise weighting term, the probability distribution of the output tokens is
\begin{equation}
P_{l,n} = \text{SoftMax}(T_{\text{cls}, l, n}), \quad Q_{l,n} = \text{SoftMax}\left( \text{Vec}(\textbf{T}_{\text{st},l,n}) \right), \quad \textbf{T}_{\text{st},l,n} = \left\{T_{st,l,m,n}\right\}_{m=1}^{M}
\end{equation}
Furthermore, in multimodal settings, each expert operates on inherently different signal distributions and modalities, making it non-trivial to determine a unified reference distribution to align with. 
This asymmetry undermines the stability and consistency of KL-based supervision.
Therefore, the inter-expert collaboration is realized by Mean Squared Error (MSE) which provides a symmetric, modality-agnostic objective for encouraging alignment across expert representations. 
The loss to measure the alignment among the experts can be denoted as
\[
L_{\text{inter}} = \frac{1}{N(N-1)} \sum_{n=1}^{N} \sum_{k=1,k\neq n}^{N} \parallel T_{\text{cls}, L, n} - T_{\text{cls}, L, k} \parallel^2_2
\]

From an optimization perspective, the final objective combines task supervision, intra-expert token refinement, and inter-expert consistency learning, the overall loss function can be written as
\[
L = L_{\text{cls}} + \alpha L_{\text{intra}} + \beta L_{\text{inter}},
\]
in which each loss term is weighted by parameters $\alpha$ and $\beta$.

Overall, MoME and HTL play complementary roles in the proposed framework. 
MoME addresses the question of how multimodal experts should be coordinated under input-varying reliability, while HTL addresses how action-relevant knowledge should be refined and transferred both within and across experts. 
Their combination forms a unified multimodal learning framework that couples adaptive expert collaboration with token-level knowledge enhancement, which is particularly suitable for fine-grained in-cabin visual analytics.

\section{Experiments}
In this section, we explain our experiments with the detail of implementation, the benchmarking results compared with related works, and ablation studies demonstrating the effectiveness of our proposed methods.

\begin{table*}
\centering
\caption{The benchmark results of MoME compared with single- and multimodal DAR baselines on Drive\&Act~\citep{martin2019daa} dataset. ``LF" denotes basic late fusion-based multimodal DAR variants implemented with different backbones}
\label{results}
\begin{tblr}{
  width=0.95\linewidth,
  colspec={X[1.8,l] X[0.8,c] X[1.1,c] X[1.1,c]},
  row{7} = {c},
  row{8} = {c},
  row{10} = {c},
  row{11} = {c},
  row{15} = {c},
  row{16} = {c},
  row{18} = {c},
  row{19} = {c},
  cell{1}{2} = {c},
  cell{1}{3} = {c},
  cell{1}{4} = {c},
  cell{2}{2} = {c},
  cell{2}{3} = {c},
  cell{2}{4} = {c},
  cell{3}{2} = {c},
  cell{3}{3} = {c},
  cell{3}{4} = {c},
  cell{4}{2} = {c},
  cell{4}{3} = {c},
  cell{4}{4} = {c},
  cell{5}{2} = {c},
  cell{5}{3} = {c},
  cell{5}{4} = {c},
  cell{6}{1} = {r=3}{},
  cell{6}{2} = {c},
  cell{6}{3} = {c},
  cell{6}{4} = {c},
  cell{9}{1} = {r=3}{},
  cell{9}{2} = {c},
  cell{9}{3} = {c},
  cell{9}{4} = {c},
  cell{12}{2} = {c},
  cell{12}{3} = {c},
  cell{12}{4} = {c},
  cell{13}{2} = {c},
  cell{13}{3} = {c},
  cell{13}{4} = {c},
  cell{14}{1} = {r=3}{},
  cell{14}{2} = {c},
  cell{14}{3} = {c},
  cell{14}{4} = {c},
  cell{17}{1} = {r=3}{},
  cell{17}{2} = {c},
  cell{17}{3} = {c},
  cell{17}{4} = {c},
  cell{20}{2} = {c},
  cell{20}{3} = {c},
  cell{20}{4} = {c},
  cell{21}{2} = {c},
  cell{21}{3} = {c},
  cell{21}{4} = {c},
  cell{22}{2} = {c},
  cell{22}{3} = {c},
  cell{22}{4} = {c},
  cell{23}{2} = {c},
  cell{23}{3} = {c},
  cell{23}{4} = {c},
  cell{24}{2} = {c},
  cell{24}{3} = {c},
  cell{24}{4} = {c},
  cell{25}{2} = {c},
  cell{25}{3} = {c},
  cell{25}{4} = {c},
  cell{26}{2} = {c},
  cell{26}{3} = {c},
  cell{26}{4} = {c},
  cell{27}{1} = {c=4}{},
  hline{1,20, 27} = {-}{0.2em},
  hline{2} = {2}{-}{},
  row{2,4,6,7,8,12,14,15,16,20,22,24,26} = {Mercury}
}
\textbf{Methods}                                                   & \textbf{Modality}      & \textbf{Mean-1 Accuracy}  & \textbf{Top-1 Accuracy} \\
Pose~\citep{martin2019daa}                                         & 3DS                    & -                         & 55.17               \\
C3D~\citep{tran2015learning}                                       & NIR                    & -                         & 43.41               \\
P3D~\citep{qiu2017learning}                                        & NIR                    & -                         & 45.32               \\
CTA-NET~\citep{wharton2021coarse}                                  & RGB                    & -                         & 65.25               \\
I3D~\citep{carreira2017quo}                                        & NIR                    & -                         & 60.80               \\
                                                                   & Depth                  & -                         & 60.52               \\
                                                                   & IR                     & -                         & 64.98               \\
TSM~\citep{lin2019tsm}                                             & Depth                  & 58.28                     & 63.76               \\
                                                                   & IR                     & 59.81                     & 67.75               \\
                                                                   & RGB                    & 62.72                     & 68.23               \\
TransDARC~\citep{peng2022transdarc}                                & RGB                    & 60.10                     & 76.17               \\
UniformerV2~\citep{li2023uniformerv2}                              & RGB                    & 61.79                     & 76.71               \\
Multifuser~\citep{wang2024multifuser}                              & Depth                  & 56.35                     & 69.21               \\
                                                                   & IR                     & 59.64                     & 72.56               \\
                                                                   & RGB                    & 61.79                     & 76.71               \\
CM$^2$-Net~\citep{wang2024cm2}                                     & Depth                  & 63.91                     & 77.15               \\
                                                                   & IR                     & 64.97                     & 80.46               \\
                                                                   & RGB                    & 69.76                     & 83.51               \\
LF-ResNet~\citep{he2016deep}                                       & I+D                    & 51.08                     & 56.43               \\
LF-TSM~\citep{lin2019tsm}                                          & I+D                    & 61.11                     & 70.31               \\
DFS~\citep{lin2024dfs}                                             & R+I                    & 62.87                     & 72.32               \\
MDBU~\citep{roitberg2022comparative}                               & Best2                  & 62.02                     & 76.91               \\
Multifuser~\citep{wang2024multifuser}                              & R+I+D                  & 70.67                     & 82.39               \\
CM$^2$-Net~\citep{wang2024cm2}                                     & R+I+D                  & 72.10                     & 83.92               \\
\textbf{Ours}                                                      & R+I+D                  & \textbf{73.82}            & \textbf{83.94}      \\
\scriptsize{3DS: 3D skeleton, N: NIR, I: IR, D: Depth, R: RGB, LF: Late Fusion} &           &                           &                     
\end{tblr}
\end{table*}

\subsection{Implementation Details}
\paragraph{\textbf{Dataset \& Metrics.}}
We conduct our experiment with Drive\&Act~\citep{martin2019daa}, which is a large-scale multimodal video dataset for driver action recognition, offering five modalities: RGB, IR, Depth, Near-InfraRed (NIR), and 3D skeleton data, collected from six camera views in real driving scenarios, and classifies activities into three levels: scenarios, fine-grained activities, and atomic actions. 
In our experiment, we focus on the practical cabin setting and commonly used setting in prior works~\citep{wang2024cm2} that focuses on classifying RGB, IR, and Depth (i.e. $N=3$) videos from the right-top view into 34 fine-grained activity classes. 
Among them, RGB offers rich appearance information, IR is more robust under challenging illumination, and Depth provides geometry-aware motion cues. 
This combination forms a representative multimodal setting for evaluating adaptive expert collaboration, it provides stable observation of the upper-body region while keeping the benchmark comparable to representative DAR studies.
We follow the three predefined splits provided by the data set for a consistent evaluation and average the results across these splits.
For performance evaluation, we use two metrics: Top-1 accuracy (Top-1 Acc.) and Mean-1 accuracy (Mean-1 Acc.). 
Top-1 Acc. measures the proportion of correct top-ranked predictions, while Mean-1 Acc. calculates the average accuracy across all classes, mitigating the effect of class imbalance.

\paragraph{\textbf{Training Configuration.}}
For each modality expert in MoME, we adopt UniformerV2-B16~\citep{li2023uniformerv2} as the video transformer backbone, which is pre-trained by Kinects-710 dataset~\citep{kay2017kinetics}.
It offers a strong transformer-based video representation model with sufficient capacity to model spatio-temporal details, making it a suitable testbed for validating whether expert collaboration and token-level learning can further improve fine-grained multimodal recognition.
For video input, we sample 8 frames, each resized to $224 \times 224$ pixels. 
These frames are then encoded into 1568 768-dimensional spatio-temporal tokens by the video transformer. 
During the training, the whole model was optimized using the AdamW optimizer \citep{loshchilov2017adamw}, with a momentum of 0.9 and a weight decay of 0.05.
The base learning rate was set to $1 \times 10^{-5}$, scaled by the number of data shards.
The loss term weight is set as $\alpha=0.01$ and $\beta=1$ according to their scale.
The learning rate followed a cosine decay~\citep{loshchilov2016sgdr}, which ended at $1 \times 10^{-7}$. 
The maximum number of training epochs was set to 100.
The models were trained on a server equipped with 2 NVIDIA GeForce RTX 4090 GPUs.

\subsection{Comparison with State-of-the-arts}
We perform a comparative analysis of our model against state-of-the-art (SOTA) driver action recognition methods. 
The compared methods cover both single-modality and multimodal DAR paradigms, including conventional video recognition backbones, fusion-based multimodal models, and recent specialized in-cabin action recognition approaches. 
This comparison allows us to examine not only whether multimodal learning is beneficial, but also whether adaptive expert collaboration provides advantages beyond existing fusion strategies.
Table. ~\ref{results} presents the Top-1 and Mean-1 Accuracy (\%) for recognizing 34 fine-grained activities in Drive\&Act Dataset, comparing our MoME with the SOTA methods. 
It can be observed that with our proposed method consistently outperforms all driver action recognition baselines in the overall comparison across all Drive\&Act splits.  
Specifically, in a multimodal setting, MoME improves Mean-1 accuracy by 1.72\% over CM$^2$-Net~\citep{wang2024cm2} as it indicates that the proposed framework better handles difficult or less frequent classes rather than only improving dominant categories. 
This observation is consistent with our motivation that expert collaboration and token-level refinement help the model capture subtle action evidence more reliably.
Although the Top-1 accuracy gain over the strongest baseline is relatively modest, the larger improvement in Mean-1 accuracy suggests that our method mainly contributes by improving class-wise robustness and reducing ambiguity in fine-grained recognition, rather than by only boosting already easy samples.
The results show that the benefit of MoME is not merely due to increasing model complexity. 
Rather, the consistent gains over representative multimodal baselines suggest that dynamically coordinating modality experts is more effective than relying on fixed fusion modules when modality usefulness changes across samples and action categories.

\subsection{Ablation Studies}
To further validate the efficacy of the MoME structure and holistic token learning strategy, we performed more detailed ablation study.

\begin{table*}[h]
\centering
\caption{Quantitative comparison across different fusion methods based on the UniformerV2~\citep{li2023uniformerv2} backbone.}
\label{tab:ab_moe}
\resizebox{\linewidth}{!}{
\begin{tblr}{
  cells = {c,m},
  cell{1}{1} = {r=2}{},
  cell{1}{2} = {c=2}{},
  cell{1}{4} = {c=2}{},
  cell{1}{6} = {c=2}{},
  cell{1}{8} = {c=2}{},
  hline{1,7} = {-}{0.2em},
  hline{3} = {-}{},
}
\textbf{Method } & \textbf{Split 0}     &                   & \textbf{Split 1}  &                   & \textbf{Split 2}  &                   & \textbf{Average}  &                      \\
\hline
                 & Mean-1 Acc.          & Top-1 Acc.        & Mean-1 Acc.       & Top-1 Acc.        & Mean-1 Acc.       & Top-1 Acc.        & Mean-1 Acc.       & Top-1 Acc.           \\
Early-Fusion     & 54.19                & 72.86             & 52.24             & 65.70             & 43.21             & 63.34             & 49.88             & 67.30                \\
Late-Fusion      & 65.34                & 84.29             & 60.16             & 77.36             & 59.25             & 75.19             & 61.58             & 78.53                \\
Multifuser~\citep{wang2024multifuser}       & 71.95                & 85.42             & 72.07             & 82.85             & 68.00             & 78.89             & 70.67             & 82.39             \\
\textbf{Ours}            & \textbf{74.32}                & \textbf{88.35}             & 7\textbf{3.54}             & \textbf{82.91}             & \textbf{73.60}             & \textbf{80.56}             & \textbf{73.82}             & \textbf{83.94}
\end{tblr}
}
\end{table*}

We first perform a detailed comparison with representative multimodal fusion methods, including traditional early and late fusion strategies as well as recent Multifuser~\citep{wang2024multifuser}, the backbone model follows the same setting.
As shown in Tab.~\ref{tab:ab_moe}, it can be observed that our method provides a noticeable improvement between different splits of the dataset, which proves the advantage of our design.

\renewcommand{\arraystretch}{1.3}
\begin{table}
    \centering
    \caption{Ablation study results of MoME design and comparison between HTL and MoDE~\citep{xie2024mode}}
    \resizebox{0.5\linewidth}{!}{
    \begin{tabular}{lccc}
        \toprule
        \textbf{Structure}      & \textbf{Strategy}  & \textbf{Mean-1 Acc.} & \textbf{Top-1 Acc.} \\
        \midrule
        MoE                   & None          & 72.14         & 87.94       \\
        MoME                  & None          & 73.48     & 88.03  \\
        MoME                  & MoDE~\citep{xie2024mode}          & 72.98     & 88.17  \\
        \textbf{MoME}                  & \textbf{HTL}           & \textbf{74.32}     & \textbf{88.35}  \\
        \bottomrule
    \end{tabular}}
    \label{tab:ab_1}
    
\end{table}

To further demonstrate the advantage of MoME structure and HTL strategy, based on the same modality expert setting, we implemented the basic MoE framework whose gating module performs expert coordination with the input sample, which comprised three linear layers to compute the weighting vector by receiving the downsampled samples.
In addition, we implement the latest MoE enhancement method, MoDE~\citep{xie2024mode} for comparison.
The result is shown in Table.~\ref{tab:ab_1}. 
Comparing MoE and MoME without auxiliary strategy shows that output-aware gating already brings measurable gain. 
This validates our design choice that routing decisions should be informed by expert outputs rather than by input features alone.
When adding expert-level enhancement, HTL performs better than MoDE. 
This suggests that class-token-level mutual distillation alone is insufficient for fine-grained DAR, where subtle motion evidence is often encoded in lower-level spatio-temporal tokens.
The final gain achieved by MoME + HTL confirms that robust multimodal recognition benefits from both adaptive expert coordination and token-level knowledge transfer.

\renewcommand{\arraystretch}{1.3}
\begin{table}[!b]
    \centering
    \caption{HTL can also be applied to signal-modality models. The single modality variant HTL - can contribute more performance improvement than other self-distillation-like methods.}
    \resizebox{0.6\linewidth}{!}{
    \begin{tabular}{lcc}
        \toprule
        \textbf{Methods}  & \textbf{Mean-1 Acc.} & \textbf{Top-1 Acc.} \\
        \midrule
        Basic Fine-tuning                              & 63.02          & 83.08 \\           
        Deep supervision~\citep{lee2015deeply}                        & 66.10          & 83.71  \\                
        Self-distillation~\citep{sultana2022self}                        & 63.96          & 83.44  \\
        \textbf{Holistic Token Learning -}                 & \textbf{69.18}          & \textbf{84.16}  \\
        \bottomrule
    \end{tabular}}
    \label{tab:ab_2}
\end{table}

\begin{figure}
    \centering
    \includegraphics[width=1\linewidth]{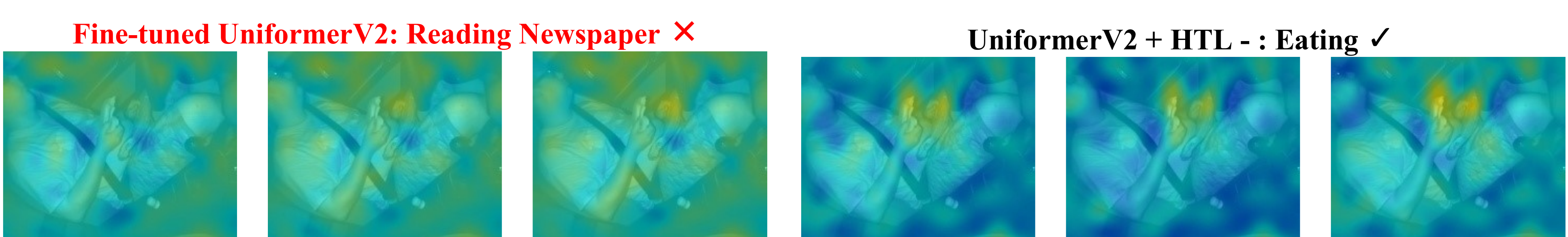}
    \caption{HTL can guide the model to concentrate on subtle spatio-temporal cues when handling challenging samples.}
    \label{fig:vis}
\end{figure}

Furthermore, our HTL can be adapted to single-modality models.
We applied our strategy on the general action recognition model and evaluated its extensibility.
From the statistics in Table.~\ref{tab:ab_2}, it can be seen that the adapted HTL- which only includes the intra-expert part still shows better performance over deep supervision~\citep{lee2015deeply} and self-distillation~\citep{sultana2022self}.
It further suggests that refining spatio-temporal tokens is more beneficial than only applying conventional intermediate supervision or class-token-based self-distillation.
And the visualized attention map in Fig.~\ref{fig:vis} demonstrates that the HTL-enhanced variant attends more selectively to subtle action-relevant regions instead of relying on diffuse background responses. 
This behavior is consistent with our design objective and also supports the interpretability of the proposed framework from a token-level evidence perspective.

\section{Conclusion}
In this paper, our proposed Mixture-of-Modality-Experts (MoME) framework, empowered by Holistic Token Learning (HTL) strategy, effectively achieves more robust multimodal driver action recognition and enhanced expert collaboration in MoE structure.
By leveraging both low-level spatio-temporal and high-level class token cues for expert specialization and inter-expert collaboration, our approach demonstrates strong adaptability to the constrained and complex in-cabin environment. 
Experiments validate its superiority in bridging the domain gap and enhancing recognition performance, which is important for enhanced human-vehicle interaction as well as more reliable intelligent transportation systems.

\printcredits

\bibliographystyle{cas-model2-names}

\bibliography{cas-refs}

\begin{thebibliography}{34}
\expandafter\ifx\csname natexlab\endcsname\relax\def\natexlab#1{#1}\fi
\providecommand{\url}[1]{\texttt{#1}}
\providecommand{\href}[2]{#2}
\providecommand{\path}[1]{#1}
\providecommand{\DOIprefix}{doi:}
\providecommand{\ArXivprefix}{arXiv:}
\providecommand{\URLprefix}{URL: }
\providecommand{\Pubmedprefix}{pmid:}
\providecommand{\doi}[1]{\href{http://dx.doi.org/#1}{\path{#1}}}
\providecommand{\Pubmed}[1]{\href{pmid:#1}{\path{#1}}}
\providecommand{\bibinfo}[2]{#2}
\ifx\xfnm\relax \def\xfnm[#1]{\unskip,\space#1}\fi
\bibitem[{Carreira and Zisserman(2017)}]{carreira2017quo}
\bibinfo{author}{Carreira, J.}, \bibinfo{author}{Zisserman, A.}, \bibinfo{year}{2017}.
\newblock \bibinfo{title}{Quo vadis, action recognition? a new model and the kinetics dataset}, in: \bibinfo{booktitle}{proceedings of the IEEE Conference on Computer Vision and Pattern Recognition}, pp. \bibinfo{pages}{6299--6308}.
\bibitem[{Chung et~al.(2022)Chung, Wu and Russakovsky}]{chung2022backgroundbias}
\bibinfo{author}{Chung, J.}, \bibinfo{author}{Wu, Y.}, \bibinfo{author}{Russakovsky, O.}, \bibinfo{year}{2022}.
\newblock \bibinfo{title}{Enabling detailed action recognition evaluation through video dataset augmentation}.
\newblock \bibinfo{journal}{Advances in Neural Information Processing Systems} \bibinfo{volume}{35}, \bibinfo{pages}{39020--39033}.
\bibitem[{Gong et~al.(2024)Gong, Wang, Zhou, Wen and Zhang}]{gong2024tfac}
\bibinfo{author}{Gong, P.}, \bibinfo{author}{Wang, P.}, \bibinfo{author}{Zhou, Y.}, \bibinfo{author}{Wen, X.}, \bibinfo{author}{Zhang, D.}, \bibinfo{year}{2024}.
\newblock \bibinfo{title}{Tfac-net: A temporal-frequential attentional convolutional network for driver drowsiness recognition with single-channel eeg}.
\newblock \bibinfo{journal}{IEEE Transactions on Intelligent Transportation Systems} .
\bibitem[{He et~al.(2016)He, Zhang, Ren and Sun}]{he2016deep}
\bibinfo{author}{He, K.}, \bibinfo{author}{Zhang, X.}, \bibinfo{author}{Ren, S.}, \bibinfo{author}{Sun, J.}, \bibinfo{year}{2016}.
\newblock \bibinfo{title}{Deep residual learning for image recognition}, in: \bibinfo{booktitle}{Proceedings of the IEEE conference on computer vision and pattern recognition}, pp. \bibinfo{pages}{770--778}.
\bibitem[{Jacobs et~al.(1991)Jacobs, Jordan, Nowlan and Hinton}]{jacobs1991moe}
\bibinfo{author}{Jacobs, R.A.}, \bibinfo{author}{Jordan, M.I.}, \bibinfo{author}{Nowlan, S.J.}, \bibinfo{author}{Hinton, G.E.}, \bibinfo{year}{1991}.
\newblock \bibinfo{title}{Adaptive mixtures of local experts}.
\newblock \bibinfo{journal}{Neural computation} \bibinfo{volume}{3}, \bibinfo{pages}{79--87}.
\bibitem[{Jiang et~al.(2021)Jiang, Hou, Yuan, Zhou, Shi, Jin, Wang and Feng}]{jiang2021tokenlabelling}
\bibinfo{author}{Jiang, Z.H.}, \bibinfo{author}{Hou, Q.}, \bibinfo{author}{Yuan, L.}, \bibinfo{author}{Zhou, D.}, \bibinfo{author}{Shi, Y.}, \bibinfo{author}{Jin, X.}, \bibinfo{author}{Wang, A.}, \bibinfo{author}{Feng, J.}, \bibinfo{year}{2021}.
\newblock \bibinfo{title}{All tokens matter: Token labeling for training better vision transformers}.
\newblock \bibinfo{journal}{Advances in Neural Information Processing Systems} \bibinfo{volume}{34}, \bibinfo{pages}{18590--18602}.
\bibitem[{Kay et~al.(2017)Kay, Carreira, Simonyan, Zhang, Hillier, Vijayanarasimhan, Viola, Green, Back, Natsev et~al.}]{kay2017kinetics}
\bibinfo{author}{Kay, W.}, \bibinfo{author}{Carreira, J.}, \bibinfo{author}{Simonyan, K.}, \bibinfo{author}{Zhang, B.}, \bibinfo{author}{Hillier, C.}, \bibinfo{author}{Vijayanarasimhan, S.}, \bibinfo{author}{Viola, F.}, \bibinfo{author}{Green, T.}, \bibinfo{author}{Back, T.}, \bibinfo{author}{Natsev, P.}, et~al., \bibinfo{year}{2017}.
\newblock \bibinfo{title}{The kinetics human action video dataset}.
\newblock \bibinfo{journal}{arXiv preprint arXiv:1705.06950} .
\bibitem[{Kuang et~al.(2023)Kuang, Li, Li, Zhang and Wu}]{kuang2023mifi}
\bibinfo{author}{Kuang, J.}, \bibinfo{author}{Li, W.}, \bibinfo{author}{Li, F.}, \bibinfo{author}{Zhang, J.}, \bibinfo{author}{Wu, Z.}, \bibinfo{year}{2023}.
\newblock \bibinfo{title}{Mifi: Multi-camera feature integration for robust 3d distracted driver activity recognition}.
\newblock \bibinfo{journal}{IEEE Transactions on Intelligent Transportation Systems} .
\bibitem[{Lee et~al.(2015)Lee, Xie, Gallagher, Zhang and Tu}]{lee2015deeply}
\bibinfo{author}{Lee, C.Y.}, \bibinfo{author}{Xie, S.}, \bibinfo{author}{Gallagher, P.}, \bibinfo{author}{Zhang, Z.}, \bibinfo{author}{Tu, Z.}, \bibinfo{year}{2015}.
\newblock \bibinfo{title}{Deeply-supervised nets}, in: \bibinfo{booktitle}{Artificial intelligence and statistics}, \bibinfo{organization}{Pmlr}. pp. \bibinfo{pages}{562--570}.
\bibitem[{Li et~al.(2023a)Li, Wang, He, Li, Wang, Wang and Qiao}]{li2023uniformerv2}
\bibinfo{author}{Li, K.}, \bibinfo{author}{Wang, Y.}, \bibinfo{author}{He, Y.}, \bibinfo{author}{Li, Y.}, \bibinfo{author}{Wang, Y.}, \bibinfo{author}{Wang, L.}, \bibinfo{author}{Qiao, Y.}, \bibinfo{year}{2023}a.
\newblock \bibinfo{title}{Uniformerv2: Unlocking the potential of image vits for video understanding}, in: \bibinfo{booktitle}{Proceedings of the IEEE/CVF International Conference on Computer Vision}, pp. \bibinfo{pages}{1632--1643}.
\bibitem[{Li et~al.(2023b)Li, Wang, Zhang, Gao, Song, Liu, Li and Qiao}]{li2023uniformer}
\bibinfo{author}{Li, K.}, \bibinfo{author}{Wang, Y.}, \bibinfo{author}{Zhang, J.}, \bibinfo{author}{Gao, P.}, \bibinfo{author}{Song, G.}, \bibinfo{author}{Liu, Y.}, \bibinfo{author}{Li, H.}, \bibinfo{author}{Qiao, Y.}, \bibinfo{year}{2023}b.
\newblock \bibinfo{title}{Uniformer: Unifying convolution and self-attention for visual recognition}.
\newblock \bibinfo{journal}{IEEE Transactions on Pattern Analysis and Machine Intelligence} \bibinfo{volume}{45}, \bibinfo{pages}{12581--12600}.
\bibitem[{Lin et~al.(2024)Lin, Lee, Li, Wang, Yap, Li and Ngim}]{lin2024dfs}
\bibinfo{author}{Lin, D.}, \bibinfo{author}{Lee, P.H.Y.}, \bibinfo{author}{Li, Y.}, \bibinfo{author}{Wang, R.}, \bibinfo{author}{Yap, K.H.}, \bibinfo{author}{Li, B.}, \bibinfo{author}{Ngim, Y.S.}, \bibinfo{year}{2024}.
\newblock \bibinfo{title}{Multi-modality action recognition based on dual feature shift in vehicle cabin monitoring}, in: \bibinfo{booktitle}{ICASSP 2024-2024 IEEE International Conference on Acoustics, Speech and Signal Processing (ICASSP)}, \bibinfo{organization}{IEEE}. pp. \bibinfo{pages}{6480--6484}.
\bibitem[{Lin et~al.(2019)Lin, Gan and Han}]{lin2019tsm}
\bibinfo{author}{Lin, J.}, \bibinfo{author}{Gan, C.}, \bibinfo{author}{Han, S.}, \bibinfo{year}{2019}.
\newblock \bibinfo{title}{Tsm: Temporal shift module for efficient video understanding}, in: \bibinfo{booktitle}{Proceedings of the IEEE/CVF international conference on computer vision}, pp. \bibinfo{pages}{7083--7093}.
\bibitem[{Liu et~al.(2026)Liu, Lu, Zhang, Cai, Gao, Wang, Yap and Chau}]{liu2026accelerating}
\bibinfo{author}{Liu, T.}, \bibinfo{author}{Lu, Y.}, \bibinfo{author}{Zhang, L.}, \bibinfo{author}{Cai, C.}, \bibinfo{author}{Gao, J.}, \bibinfo{author}{Wang, Y.}, \bibinfo{author}{Yap, K.H.}, \bibinfo{author}{Chau, L.P.}, \bibinfo{year}{2026}.
\newblock \bibinfo{title}{Accelerating diffusion-based video editing via heterogeneous caching: Beyond full computing at sampled denoising timestep}.
\newblock \bibinfo{journal}{arXiv preprint arXiv:2603.24260} .
\bibitem[{Liu and Sugano(2022)}]{liu2022interactive}
\bibinfo{author}{Liu, T.}, \bibinfo{author}{Sugano, Y.}, \bibinfo{year}{2022}.
\newblock \bibinfo{title}{Interactive machine learning on edge devices with user-in-the-loop sample recommendation}.
\newblock \bibinfo{journal}{IEEE Access} \bibinfo{volume}{10}, \bibinfo{pages}{107346--107360}.
\bibitem[{Liu et~al.(2025)Liu, Wu, Cai, Wang, Yap and Chau}]{liu2025towards}
\bibinfo{author}{Liu, T.}, \bibinfo{author}{Wu, K.}, \bibinfo{author}{Cai, C.}, \bibinfo{author}{Wang, Y.}, \bibinfo{author}{Yap, K.H.}, \bibinfo{author}{Chau, L.P.}, \bibinfo{year}{2025}.
\newblock \bibinfo{title}{Towards blind bitstream-corrupted video recovery: A visual foundation model-driven framework}, in: \bibinfo{booktitle}{Proceedings of the 33rd ACM International Conference on Multimedia}, pp. \bibinfo{pages}{7949--7958}.
\bibitem[{Liu et~al.(2023)Liu, Wu, Wang, Liu, Yap and Chau}]{liu2024bitstream}
\bibinfo{author}{Liu, T.}, \bibinfo{author}{Wu, K.}, \bibinfo{author}{Wang, Y.}, \bibinfo{author}{Liu, W.}, \bibinfo{author}{Yap, K.H.}, \bibinfo{author}{Chau, L.P.}, \bibinfo{year}{2023}.
\newblock \bibinfo{title}{Bitstream-corrupted video recovery: A novel benchmark dataset and method}.
\newblock \bibinfo{journal}{Advances in Neural Information Processing Systems} \bibinfo{volume}{36}, \bibinfo{pages}{68420--68433}.
\bibitem[{Loshchilov(2017)}]{loshchilov2017adamw}
\bibinfo{author}{Loshchilov, I.}, \bibinfo{year}{2017}.
\newblock \bibinfo{title}{Decoupled weight decay regularization}.
\newblock \bibinfo{journal}{arXiv preprint arXiv:1711.05101} .
\bibitem[{Loshchilov and Hutter(2016)}]{loshchilov2016sgdr}
\bibinfo{author}{Loshchilov, I.}, \bibinfo{author}{Hutter, F.}, \bibinfo{year}{2016}.
\newblock \bibinfo{title}{Sgdr: Stochastic gradient descent with warm restarts}.
\newblock \bibinfo{journal}{arXiv preprint arXiv:1608.03983} .
\bibitem[{Lv et~al.(2022)Lv, Nian, Xu and Song}]{lv2022compact}
\bibinfo{author}{Lv, C.}, \bibinfo{author}{Nian, J.}, \bibinfo{author}{Xu, Y.}, \bibinfo{author}{Song, B.}, \bibinfo{year}{2022}.
\newblock \bibinfo{title}{Compact vehicle driver fatigue recognition technology based on eeg signal}.
\newblock \bibinfo{journal}{IEEE Transactions on Intelligent Transportation Systems} \bibinfo{volume}{23}, \bibinfo{pages}{19753--19759}.
\newblock \DOIprefix\doi{10.1109/TITS.2021.3119354}.
\bibitem[{Ma et~al.(2017)Ma, Chau and Yap}]{ma2017depth}
\bibinfo{author}{Ma, X.}, \bibinfo{author}{Chau, L.P.}, \bibinfo{author}{Yap, K.H.}, \bibinfo{year}{2017}.
\newblock \bibinfo{title}{Depth video-based two-stream convolutional neural networks for driver fatigue detection}, in: \bibinfo{booktitle}{2017 International Conference on Orange Technologies (ICOT)}, \bibinfo{organization}{IEEE}. pp. \bibinfo{pages}{155--158}.
\bibitem[{Ma et~al.(2019)Ma, Chau, Yap and Ping}]{ma2019conv}
\bibinfo{author}{Ma, X.}, \bibinfo{author}{Chau, L.P.}, \bibinfo{author}{Yap, K.H.}, \bibinfo{author}{Ping, G.}, \bibinfo{year}{2019}.
\newblock \bibinfo{title}{Convolutional three-stream network fusion for driver fatigue detection from infrared videos}, in: \bibinfo{booktitle}{2019 IEEE International Symposium on Circuits and Systems (ISCAS)}, \bibinfo{organization}{IEEE}. pp. \bibinfo{pages}{1--5}.
\bibitem[{Martin et~al.(2019)Martin, Roitberg, Haurilet, Horne, Rei{\ss}, Voit and Stiefelhagen}]{martin2019daa}
\bibinfo{author}{Martin, M.}, \bibinfo{author}{Roitberg, A.}, \bibinfo{author}{Haurilet, M.}, \bibinfo{author}{Horne, M.}, \bibinfo{author}{Rei{\ss}, S.}, \bibinfo{author}{Voit, M.}, \bibinfo{author}{Stiefelhagen, R.}, \bibinfo{year}{2019}.
\newblock \bibinfo{title}{Drive\&act: A multi-modal dataset for fine-grained driver behavior recognition in autonomous vehicles}, in: \bibinfo{booktitle}{Proceedings of the IEEE/CVF International Conference on Computer Vision}, pp. \bibinfo{pages}{2801--2810}.
\bibitem[{Peng et~al.(2022)Peng, Roitberg, Yang, Zhang and Stiefelhagen}]{peng2022transdarc}
\bibinfo{author}{Peng, K.}, \bibinfo{author}{Roitberg, A.}, \bibinfo{author}{Yang, K.}, \bibinfo{author}{Zhang, J.}, \bibinfo{author}{Stiefelhagen, R.}, \bibinfo{year}{2022}.
\newblock \bibinfo{title}{Transdarc: Transformer-based driver activity recognition with latent space feature calibration}, in: \bibinfo{booktitle}{2022 IEEE/RSJ International Conference on Intelligent Robots and Systems (IROS)}, \bibinfo{organization}{IEEE}. pp. \bibinfo{pages}{278--285}.
\bibitem[{Qiu et~al.(2017)Qiu, Yao and Mei}]{qiu2017learning}
\bibinfo{author}{Qiu, Z.}, \bibinfo{author}{Yao, T.}, \bibinfo{author}{Mei, T.}, \bibinfo{year}{2017}.
\newblock \bibinfo{title}{Learning spatio-temporal representation with pseudo-3d residual networks}, in: \bibinfo{booktitle}{proceedings of the IEEE International Conference on Computer Vision}, pp. \bibinfo{pages}{5533--5541}.
\bibitem[{Roitberg et~al.(2022)Roitberg, Peng, Marinov, Seibold, Schneider and Stiefelhagen}]{roitberg2022comparative}
\bibinfo{author}{Roitberg, A.}, \bibinfo{author}{Peng, K.}, \bibinfo{author}{Marinov, Z.}, \bibinfo{author}{Seibold, C.}, \bibinfo{author}{Schneider, D.}, \bibinfo{author}{Stiefelhagen, R.}, \bibinfo{year}{2022}.
\newblock \bibinfo{title}{A comparative analysis of decision-level fusion for multimodal driver behaviour understanding}, in: \bibinfo{booktitle}{2022 IEEE intelligent vehicles symposium (IV)}, \bibinfo{organization}{IEEE}. pp. \bibinfo{pages}{1438--1444}.
\bibitem[{Sultana et~al.(2022)Sultana, Naseer, Khan, Khan and Khan}]{sultana2022self}
\bibinfo{author}{Sultana, M.}, \bibinfo{author}{Naseer, M.}, \bibinfo{author}{Khan, M.H.}, \bibinfo{author}{Khan, S.}, \bibinfo{author}{Khan, F.S.}, \bibinfo{year}{2022}.
\newblock \bibinfo{title}{Self-distilled vision transformer for domain generalization}, in: \bibinfo{booktitle}{Proceedings of the Asian conference on computer vision}, pp. \bibinfo{pages}{3068--3085}.
\bibitem[{Tan et~al.(2021)Tan, Ni, Liu, Zhang, Wu, Wang and Zeng}]{tan2021bidirectional}
\bibinfo{author}{Tan, M.}, \bibinfo{author}{Ni, G.}, \bibinfo{author}{Liu, X.}, \bibinfo{author}{Zhang, S.}, \bibinfo{author}{Wu, X.}, \bibinfo{author}{Wang, Y.}, \bibinfo{author}{Zeng, R.}, \bibinfo{year}{2021}.
\newblock \bibinfo{title}{Bidirectional posture-appearance interaction network for driver behavior recognition}.
\newblock \bibinfo{journal}{IEEE Transactions on Intelligent Transportation Systems} \bibinfo{volume}{23}, \bibinfo{pages}{13242--13254}.
\bibitem[{Tran et~al.(2015)Tran, Bourdev, Fergus, Torresani and Paluri}]{tran2015learning}
\bibinfo{author}{Tran, D.}, \bibinfo{author}{Bourdev, L.}, \bibinfo{author}{Fergus, R.}, \bibinfo{author}{Torresani, L.}, \bibinfo{author}{Paluri, M.}, \bibinfo{year}{2015}.
\newblock \bibinfo{title}{Learning spatiotemporal features with 3d convolutional networks}, in: \bibinfo{booktitle}{Proceedings of the IEEE international conference on computer vision}, pp. \bibinfo{pages}{4489--4497}.
\bibitem[{Wang et~al.(2024a)Wang, Cai, Wang, Gao, Lin, Liu and Yap}]{wang2024cm2}
\bibinfo{author}{Wang, R.}, \bibinfo{author}{Cai, C.}, \bibinfo{author}{Wang, W.}, \bibinfo{author}{Gao, J.}, \bibinfo{author}{Lin, D.}, \bibinfo{author}{Liu, W.}, \bibinfo{author}{Yap, K.H.}, \bibinfo{year}{2024}a.
\newblock \bibinfo{title}{Cm2-net: Continual cross-modal mapping network for driver action recognition}.
\newblock \bibinfo{journal}{arXiv preprint arXiv:2406.11340} .
\bibitem[{Wang et~al.(2024b)Wang, Wang, Gao, Lin, Yap and Li}]{wang2024multifuser}
\bibinfo{author}{Wang, R.}, \bibinfo{author}{Wang, W.}, \bibinfo{author}{Gao, J.}, \bibinfo{author}{Lin, D.}, \bibinfo{author}{Yap, K.H.}, \bibinfo{author}{Li, B.}, \bibinfo{year}{2024}b.
\newblock \bibinfo{title}{Multifuser: Multimodal fusion transformer for enhanced driver action recognition}.
\newblock \bibinfo{journal}{arXiv preprint arXiv:2408.01766} .
\bibitem[{Wharton et~al.(2021)Wharton, Behera, Liu and Bessis}]{wharton2021coarse}
\bibinfo{author}{Wharton, Z.}, \bibinfo{author}{Behera, A.}, \bibinfo{author}{Liu, Y.}, \bibinfo{author}{Bessis, N.}, \bibinfo{year}{2021}.
\newblock \bibinfo{title}{Coarse temporal attention network (cta-net) for driver's activity recognition}, in: \bibinfo{booktitle}{Proceedings of the IEEE/CVF winter conference on applications of computer vision}, pp. \bibinfo{pages}{1279--1289}.
\bibitem[{Xie et~al.(2024)Xie, Zhang, Zhuang, Shi, Liu, Gu and Zhang}]{xie2024mode}
\bibinfo{author}{Xie, Z.}, \bibinfo{author}{Zhang, Y.}, \bibinfo{author}{Zhuang, C.}, \bibinfo{author}{Shi, Q.}, \bibinfo{author}{Liu, Z.}, \bibinfo{author}{Gu, J.}, \bibinfo{author}{Zhang, G.}, \bibinfo{year}{2024}.
\newblock \bibinfo{title}{Mode: A mixture-of-experts model with mutual distillation among the experts}, in: \bibinfo{booktitle}{Proceedings of the AAAI Conference on Artificial Intelligence}, pp. \bibinfo{pages}{16067--16075}.
\bibitem[{Yang et~al.(2020)Yang, Liu, Min, Yang and Xiong}]{yang2020yawn}
\bibinfo{author}{Yang, H.}, \bibinfo{author}{Liu, L.}, \bibinfo{author}{Min, W.}, \bibinfo{author}{Yang, X.}, \bibinfo{author}{Xiong, X.}, \bibinfo{year}{2020}.
\newblock \bibinfo{title}{Driver yawning detection based on subtle facial action recognition}.
\newblock \bibinfo{journal}{IEEE Transactions on Multimedia} \bibinfo{volume}{23}, \bibinfo{pages}{572--583}.

\end{thebibliography}






\end{document}